\documentclass{ecai}

\usepackage{graphicx}
\usepackage{latexsym}
\usepackage{microtype}
\usepackage[capitalise,noabbrev]{cleveref}

\paperid{959}

\begin{document}

\begin{frontmatter}

\title{Bridging the Transparency Gap: \\ What Can Explainable AI Learn From the AI Act?}

\author[A]{\fnms{Balint}~\snm{Gyevnar}\thanks{Corresponding Author. Email: balint.gyevnar@ed.ac.uk}}
\author[A]{\fnms{Nick}~\snm{Ferguson}}
\author[B]{\fnms{Burkhard}~\snm{Schafer}}

\address[A]{School of Informatics, University of Edinburgh}
\address[B]{Edinburgh Law School, University of Edinburgh}

\begin{abstract}
    The European Union has proposed the Artificial Intelligence Act which introduces detailed requirements of transparency for AI systems.
    Many of these requirements can be addressed by the field of explainable AI (XAI), however, there is a fundamental difference between XAI and the Act regarding what transparency is.
    The Act views transparency as a means that supports wider values, such as accountability, human rights, and sustainable innovation.
    In contrast, XAI views transparency narrowly as an end in itself, focusing on explaining complex algorithmic properties without considering the socio-technical context.
    We call this difference the ``transparency gap''. 
    Failing to address the transparency gap, XAI risks leaving a range of transparency issues unaddressed.
    To begin to bridge this gap, we overview and clarify the terminology of how XAI and European regulation -- the Act and the related General Data Protection Regulation (GDPR) -- view basic definitions of transparency.
    By comparing the disparate views of XAI and regulation, we arrive at four axes where practical work could bridge the transparency gap: defining the scope of transparency, clarifying the legal status of XAI, addressing issues with conformity assessment, and building explainability for datasets.

\end{abstract}

 \end{frontmatter}

\section{Introduction}\label{sec:intro}
Artificial Intelligence (AI) systems are an increasing presence in twenty-first-century life and are now available to most, from commercial users to non-technical enthusiasts.
While public-facing demonstrations of the capability of AI may lead to the perception that AI is an emerging field, the use of AI systems is already widespread.
Facial recognition, recommender systems, and medical diagnoses are applications which have been utilising AI for many years.
However, as we see black-box AI increasingly deployed in safety-critical applications, issues relating to their deployment in society have arisen.

Recognising the urgent need to address such concerns, regulators have been drawing up proposals to tackle the challenges of the ``AI-era''.
In the USA, the White House has introduced the ``Blueprint for an AI Bill of Rights'', proposing five main areas of regulation for AI~\cite{thewhitehouseBlueprintAIBill2022}.
The UK has outlined its own pro-innovation approach to AI regulation and is working on a ``National AI Strategy''~\cite{dorriesEstablishingProinnovationApproach2022}.
In the EU, an ambitious regulation called the \textit{AI Act} is under consideration~\cite{europeancommissionProposalArtificialIntelligence2021}.
The Act places requirements on AI providers (developers of AI systems) relating to transparency and explainability. 
It takes a proportional risk-based approach to defining these requirements and proposes comprehensive conformity assessment conditions.

Scientists also acknowledge the need for transparent and trustworthy AI~\cite{aihlegEthicsGuidelinesTrustworthy2019,kaurTrustworthyArtificialIntelligence2022,stanfordAIIndexReport2023} that respect the law, ethics, and social considerations, but which are also robust in the real-world.
Researchers regard explanations as essential to transparency in human-AI interaction and explainable AI (XAI) now focuses to a large extent on achieving human-aligned AI via social explanations~\cite{millerExplanationArtificialIntelligence2019,saralajewHumanCentricAssessmentFramework2022,sovranoMetricsExplainabilityEuropean2022}.

However, the definitions, scope, and purpose of transparency in regulations are not in agreement with how technological approaches understand them~\cite{nanniniExplainabilityAIPolicies2023}, let alone how they translate into more trustworthy systems~\cite{lauxTrustworthyArtificialIntelligence2023}.
XAI views transparency merely as an algorithmic property that offers practical solutions but through a limited, technology-focused scope~\cite{liaoConnectingAlgorithmicResearch2022, millerExplainableAIDead2023}.
By contrast, the law treats transparency as a quality of complex socio-technical interactions between the AI and its users, developers, owners, and wider society. 
We call this mismatch of views the \textbf{``transparency gap''}.
Transparency in the view of the law is not a goal in itself, but a \textit{means} that is needed to promote a range of very different values. 
As a consequence regulation in the EU differentiates among various forms of technical, enabling, and protective transparency~\cite{hackerVarietiesAIExplanations2022}, which may be amenable only in varying degrees to computational solutions.
XAI by contrast discusses concepts of algorithmic transparency -- e.g., black box or interpretable systems -- as formal properties of a computer system in isolation~\cite{schwalbeComprehensiveTaxonomyExplainable2022}.
In this view, achieving transparency is an \textit{end} in itself, necessary by virtue of the complexity of AI systems.
Thus, the transparency gap is also the difference in viewing transparency as a means or an end.

The transparency gap is also one of the reasons why many of the legal requirements are being criticised by writers from the computer science community as being ineffective, overreaching, or technically infeasible~\cite{laion.aiCallProtectOpenSource2023, vealeDemystifyingDraftEU2021}.
The mistake here is to assume that the broad and ambitious transparency requirements that the Act lays out are engineering instructions, and can be addressed solely through appropriate design decisions. 
A much better understanding however is to ask which design decisions facilitate, and which ones hinder, the type of transparency aspiration that the Act aims for, without equating algorithmic explainability with legal transparency outright.

This type of mismatch is not a new phenomenon.
System theory teaches us that law, like all social systems, is cognitively open but normatively closed~\cite{teubnerLawAutopoieticSystem1993}. 
This means, for instance, that the concept of ``causation'' -- in science understood as a purely descriptive term used for the explanation of observations -- is ``normatively constrained''~\cite{Hart1959-HARCIT-4} so that not every causation in the sense of science is also causation in law, turning scientific explanations into legal justifications.
This is sometimes experienced as a misunderstanding: if only legislators understood technology better, many scientists might feel, they could legislate more appropriately, and for scientists that means in a way so that the legal norms directly translate into system requirements.
It is certainly true that some technology regulation is deeply misguided due to insufficient technological competency by the legislator (e.g.,~\cite{curtissComputerFraudAbuse2016}). 
However, much more common is a mutual and indeed necessary or inevitable misunderstanding. 
The law cannot but distort the technical conceptions that originate in science discourse if it wants to stay true to its own logic~\cite{teubnerBreakingFramesGlobal1997}. 
We argue instead that good XAI should not try to usurp the role of the law, or “solve” the legal problem of transparency, but can nonetheless anticipate how the law will (mis)understand its concepts, and in this way find new approaches to assist the legal system in achieving its objective, and bridge the transparency gap.

To this end, we first clarify and compare the terminology of how XAI and European regulation -- the GDPR and the Act -- view basic definitions of transparency~(\cref{sec:xai,sec:law}), establishing the two disparate perspectives. 
Based on this discussion, we then identify four axes (\cref{sec:misalignment}) along which cross-disciplinary work should be placed to begin to bridge the transparency gap:

\begin{enumerate}
  \item \textbf{Define transparency (\cref{ssec:misalignment:transparency})}. To increase legal certainty and to inform the design of XAI systems basic notions such as transparency, interpretability, and explainability should be clearly defined and scoped.
  \item \textbf{Clarify the legal status of XAI (\cref{ssec:misalignment:definitions})}. XAI methods are often based on methods which fit the legal definition of AI system in the Act. An AI system and an XAI tool should be considered as one unit, but the Act may treat the systems separately, conflating the transparency requirements and their consequences.
  \item \textbf{Conformity assessments (\cref{ssec:misalignment:ca})}. The Act lacks guidance on the process of certifying algorithmic transparency, which raises the question of how XAI systems can be certified for transparency while leaving open the possibility of self-regulation by providers. %
  \item \textbf{Explainable data (\cref{ssec:misalignment:data})}. The Act defines strong requirements for data quality and governance. XAI has so far neglected data transparency but should extend to explaining the effects of inherent properties of datasets on the functioning of AI systems.
\end{enumerate}

\section{Transparency and Explainable AI}\label{sec:xai}

Before discussing XAI, it is important to understand from where current concerns about opaque AI systems originate.
The first AI systems were based on symbolic logic, where knowledge about the world is represented using mathematical symbols.
The first commercially viable AI, known as \textit{expert systems}, were built on this paradigm~\cite{russelArtificialIntelligenceModern2022}.
The representation of knowledge in this manner lends an inherent interpretability in the form of a causal chain of reasoning, which aligns with human cognition making them intelligible to people~\cite{malleHowPeopleExplain1999,millerExplanationArtificialIntelligence2019}. 

In contrast, deep learning models that dominate today are not built on such tangible representations of data.
These models, usually based on \textit{neural networks}, consist of many millions of parameters.
The values of these parameters are \textit{learned}, requiring vast amounts of data, and serve to mathematically transform the input data to an output prediction or classification.
Fundamentally, this family of approaches lacks intrinsic interpretability due to the built-in parameterisation and abstraction.
While these models are highly performant, public opinion has shifted towards expressing fundamental concerns about their social, economic, ethical, and political effects~\cite{kelleyExcitingUsefulWorrying2021,schoefferThereNotEnough2022} often attributed to a loss of autonomy and a socio-technical blindness exacerbated by unclear scientific public discourse~\cite{johnsonReframingAIDiscourse2017}.

Recognising that a lack of transparency is diminishing trust in AI systems, methods that explain AI models were developed, forming the field of XAI.
Most surveys of the field~\cite{burkartSurveyExplainabilitySupervised2021, guidottiSurveyMethodsExplaining2018,mittelstadtExplainingExplanationsAI2019a,mohseniMultidisciplinarySurveyFramework2021,stepinSurveyContrastiveCounterfactual2021} define the methods of XAI as self-explanatory systems that reveal the reasoning behind the outputs of AI models.
In addition, we give a practical lower-level definition in terms of four attributes to help regulators pin down what XAI is.
An explainable AI system should:

\begin{enumerate}
    \item explain the output of an AI system;
    \item using partially or fully automated methods;
    \item to clearly defined stakeholders;
    \item in a relevant and accurate manner.
\end{enumerate}

\subsection{Explaining systems' outputs}\label{ssec:xai:output}

Different AI systems have different inherent properties that make their outputs more or less suited for explanations.
There is a significant amount of debate in XAI regarding the meanings of interpretability, explainability, justification, and transparency, and their relationships.
It is important to clarify these terms as they are not interchangeable, and their interpretations lead to different understandings of the responsibilities and capabilities of AI systems.
Issues arising from this in a legal context are further discussed in \cref{ssec:misalignment:transparency}.

Miller~\cite{millerExplanationArtificialIntelligence2019} equates \textit{interpretability} with explainability. 
However, others argue, and we support this view, that while interpretability and explainability are connected, they are distinct properties.
According to Molnar~\cite{molnarInterpretableMachineLearning2023}, what makes some AI systems interpretable, and with that inherently human-understandable, is their low complexity. 
A reasonably skilled human user can understand the output of such a system, and how it was derived from the input, even in the absence of an explanation generated by the AI. 
For example, a simple linear regression function used to predict the future value of the population of a country based on historical data is interpretable: a reasonably skilled user understands how the data informs the outcome. 

In contrast, \textit{explainability} is the property of any AI system whose output comes with an automatically generated output that has the syntactic form of an explanation. 
It is, in other words, the ability of an AI system to relevantly communicate the reasoning behind its decision-making process.
A linear regressor on its own is not explainable: showing model weights to a layperson will likely mean nothing to them.
However, matched with a suitable XAI technique a linear regressor can intelligibly communicate the relevant weights which affected its decision and thus becomes explainable.

A \textit{justification} gives a teleological rather than a mechanistic explanation. 
In Miller's words, `it explains why a decision is good, but does not necessarily aim to give an explanation of the actual decision making process'~\cite[p8]{millerExplanationArtificialIntelligence2019}.
A self-driving car might justify stopping for a pedestrian as the ``lawfully correct and safe action''  without explaining the mechanisms that transformed the input into that decision.

Finally, if we focus solely on the algorithmic properties of an AI system, then \textit{transparency} becomes the same concept as interpretability~\cite{angelovExplainableArtificialIntelligence2021,mittelstadtExplainingExplanationsAI2019a}.
However, as shown in \cref{ssec:misalignment:transparency}, this interpretation is too restrictive because it clashes with a broader vision of transparency, such as the top-down view advocated by the Act.

\subsection{Methods of XAI}\label{sec:xai:methods}

Current XAI methods include, among others, analysis of feature importance~\cite{ribeiroWhyShouldTrust2016}, saliency maps~\cite{petsiukBlackBoxExplanationObject2021}, counterfactual methods~\cite{gyevnarCausalExplanationsStochastic2023}, recognising textual entailment~\cite{majumderKnowledgeGroundedSelfRationalizationExtractive2022}, and knowledge distillation~\cite{chenCPKDConceptsProberGuidedKnowledge2021}.
Many XAI methods are themselves AI systems (e.g., natural language inference~\cite{camburuESNLINaturalLanguage2018}) by the definition of an AI system in the Act which raises further legal questions, discussed further in~\cref{ssec:misalignment:definitions}.

Researchers often categorise AI systems based on interpretability~\cite{burkartSurveyExplainabilitySupervised2021,rudinInterpretableMachineLearning2022}, contrasting non-interpretable black box systems such as neural networks against interpretable white box systems such as decision trees.
Interpretability has a crucial influence on the design choices of XAI methods: using white-box models, we can more easily guarantee verifiability and causality.
However, this usually comes at the cost of expressiveness and accuracy~\cite{carvalhoMachineLearningInterpretability2019}.
Both in regulation and technology, careful balancing is needed so that interpretability and accuracy are present at the desired levels.

Explainability also plays a key role in determining the kind of transparency XAI systems can offer.
\textit{Ante-hoc} explanations are generated directly from the internal representations and processes of white box systems, while \textit{post-hoc} explanations are inferred from an output after a decision was made.
Thus, ante-hoc explanations are truthful to the decision process by design.
Post-hoc explanations may distort the causality underlying the model's decision process and require more effort to generate, but they apply to both white and black box systems.

Finally, an XAI system can also be \textit{model-agnostic}, meaning that it can be applied to explain many AI algorithms, or it can be \textit{model-specific}, meaning that it applies to one specific AI algorithm.
While model-agnostic XAI offers off-the-shelf explainability for AI systems and, thus, can provide significant savings in resources, it raises issues of liability when the system is not sufficiently certified.
Is the due diligence of the AI provider called into question due to their selection of an unsuitable XAI system even when the underlying AI system functions properly?
This raises issues around the certification process, which we tackle in \cref{ssec:misalignment:ca}.

\subsection{Stakeholders}\label{sec:xai:stakeholder}

We must also consider how transparency can be achieved for different stakeholders of AI.
To this end, Langer et al.~\cite{langerWhatWeWant2021} have given a taxonomy of XAI from the perspective of stakeholders. 
They consider a feedback loop between the explanation process and the stakeholders based on four groups: developers, users, deployers, and affected parties.
Their categorisation aligns with the definitions of the Act, which considers similar stakeholder groups.
For example, Article 3 of the Act gives legal definitions of the provider, user, importer, etc. 

Additionally, Mohseni et al.~\cite{mohseniMultidisciplinarySurveyFramework2021} suggest three distinct end-users for XAI\@: \textit{AI novices}, \textit{data experts}, and \textit{AI experts}.
The first category is of utmost importance, as it relates to end-users with a negligible amount of expertise on how the system works. 
Their concerns include, among others, bias, privacy, and trust, which are issues that regulators are addressing in the Act, and which increased transparency is supposed to alleviate.
Multi-modal explanations, natural language communication, conversational agents, and cognitive modelling are some of the tools that are popular for addressing concerns of AI novices~\cite{ramchurnTrustworthyHumanAIPartnerships2021}.
Much theoretical and practical progress has been made in developing social XAI that addresses these concerns, but additional stakeholder-focused interdisciplinary research is sorely needed.

\subsection{Accuracy and Relevancy}

Finally, the question of evaluating the quality of generated explanations remains, especially its effects on the perceived transparency of the AI system.
Here, we must specify the desired characteristics of performance metrics, which will depend on the various stakeholders.
For example, for AI novices, measures of trust and intelligibility will be essential, while for providers, we might expect objective correctness to be more important.
There has been work on creating metrics for explanation performance evaluation~\cite{hoffmanMetricsExplainableAI2019}, however, the design of metrics that clearly show compliance with regulation is a new challenge~\cite{sovranoMetricsExplainabilityEuropean2022}.

Metrics are also essential as different XAI methods applied to the same data can produce very different explanations~\cite{renardUnderstandingPredictionDiscrepancies2021}. 
This means it is difficult to establish trustworthy \textit{baselines} unless we define and measure clearly what aspects of explanations are important, and for whom.
Comparisons to baselines are essential for demonstrating the abilities of any XAI system and regulatory requirements on explainability may well demand such comparisons to show conformity.

\section{Transparency and the Law}\label{sec:law}
As we saw, XAI delivers algorithmic transparency, but its approaches are focused on technical aspects, despite recent calls for a stakeholder-directed approach focusing on trustworthiness.

This leads us back to the transparency gap.
In the legal context, transparency is most often seen as a means to achieve broader goals, most importantly here, algorithmic accountability~\cite{hildebrandtDawnCriticalTransparency2012,kaminskiRightExplanationExplained2018}, yet \textit{not} necessarily trustworthiness.
While a trustworthy system gives the user good reasons to accept the output of the system as correct, an accountable system allows them to allocate blame appropriately if the outcome turns out to be incorrect.
In considering accountability, laws often fulfil a dual function.
They try to prevent harm from occurring, but they also allocate responsibility if harm does incur nonetheless. 
While both objectives require transparency, they may require different conceptions of transparency to fulfil these objectives.
In this context, the transparency gap is not just the inter-disciplinary means-end friction we saw in the introduction, but also an \textit{intra}-disciplinary friction as XAI figures out to what end it is building transparent systems.

To bridge the transparency gap, we need to understand the transparency requirements in the AI Act, and here it helps to put them in a historical context. 
A ``right to explanation'' for automated decisions was first trialled in the landmark data protection act of the EU, the General Data Protection Regulation (GDPR)~\cite{europeanparliamentRegulationEU20162016}.
By tracing the genealogy of the concept of explainability in the AI Act to its predecessor in the GDPR, and identifying both continuities and differences in the legislative language, we can get a better sense of the scope and limits of this provision and how they relate to the transparency gap.

\subsection{The Right to an Explanation}\label{sec:law:right}

Goodman and Flaxman~\cite{goodmanEuropeanUnionRegulations2017} first suggested that one can derive a \textit{``right to explanation''} from Article 22(3) and Articles 13--15 GDPR\@, whereby a data subject has the right to `express his or her point of view and to contest the decision' which is `based solely on automated processing', and to obtain `meaningful information about the logic involved' in the processing of personal data.
This requirement for an explanation also appears explicitly in the non-binding Recital 71.\footnote{Recitals are part of the preamble to a treaty that `articulate shared assumptions, goals and explanations concerning the treaty'~\cite[p86]{hildebrandtLawComputerScientists2020}.}

Kaminski~\cite{kaminskiRightExplanationExplained2018} supports this view by arguing that the `GDPR establishes multiple layers of transparency' in which `there is a clear relationship between the individual rights the GDPR establishes---contestation, correction, and erasure---and the kind of individualized transparency it requires.'
Malgieri and Comandé~\cite{malgieriWhyRightLegibility2017} further refined the ``right to explanation'' by combining the rights to transparency and comprehensibility to distinguish between different levels of information.
Much of these arguments are substantiated by the guidelines of the former data protection advisory board of the EU, Article 29 Data Protection Working Party (WP29)~\cite{article29dataprotectionworkingpartyGuidelinesAutomatedIndividual2018}, which include a discussion of a ``right to be informed'' and notice mechanisms for automated decision-making.
Additionally, Casey et al.~\cite{caseyRethinkingExplainableMachines2018} cite the EU data protection authorities to argue that algorithmic auditing and ``data protection by design'' methodologies codified by the GDPR are really what substantiate a ``right to explanation''.
Furthermore, Winikoff and Sardelić~\cite{winikoffArtificialIntelligenceRight2021} suggested a ``right to explanation'' could be derived from human rights in specific cases, for example, discrimination due to machine bias, which is a recurring issue of automated profiling.

However, both the existence and the utility of such a ``right to explanation'' have been called into question~\cite{wachterWhyRightExplanation2017}.
Importantly, it is argued that there are both legal and functional issues with such a right, and the intentionally vague phrasing of the GDPR makes the interpretation of Article 22 challenging.
For example, it is unclear whether the GDPR requires an ex-ante or an ex-post explanation.                                          
This was acknowledged by WP29~\cite{article29dataprotectionworkingpartyGuidelinesAutomatedIndividual2018}, which stated that explanations need not be `complex mathematical explanation[s] about how algorithms or machine-learning work', instead they should be `clear and comprehensive ways to deliver information to the data subject'.

Edwards and Veale~\cite{edwardsEnslavingAlgorithmRight2018} also warn against the ``illusion of a legal right''. 
Given the inherent technological difficulties in generating helpful explanations, the danger is that the Act has the unintended consequence of permitting low-quality automated decisions as long as the affected party gets some form of explanation, even if they won't be able to in practice use this explanation to challenge the outcome. 
Instead, Edwards and Veale suggest several actionable routes to ensure a ``right to better decisions'', e.g., via data protection impact assessments.
The arguments for the ``illusion of a legal right'' are further supported by Bayamlıoğlu~\cite{bayamliogluRightContestAutomated2022} who emphasise that it is a ``right to contest'' in Article 22(3) that should drive the discussion around tangible ways to achieve transparency.
In this view, transparency is no longer an end in itself, but a means to achieve effective contestation, and should be evaluated by the contribution they make to this aim.
As we will see, this is a view that the Act takes to a great extent.

XAI researchers have continually used the ``right to explanation'' to justify their design choices (e.g., in \cite{stepinSurveyContrastiveCounterfactual2021,atakishiyevExplainableArtificialIntelligence2021,hoffmanMetricsExplainableAI2019,guidottiSurveyMethodsExplaining2018}), but they should remember that existence and scope of a new algorithm-centric ``right to explanation'' under the GDPR remain contested and so far no case law exists on its interpretation in this aspect.
However, recently the Court of Justice of the European Union (CJEU) has advanced the debate in the beginnings of a landmark case which might favour the interpretation that a ``right to explanation'' exists~\cite{thecourtofjusticeoftheeurpeanunionRequestPreliminaryRuling}.

Even if such an interpretation is upheld, the transparency gap remains, as XAI would still regard the law as a justification for its own motivations.
A better way to bridge the transparency gap may come from Jongepier and Keymolen~\cite{jongepierExplanationAgencyExploring2022}.
Regardless of the involvement of machines, there is often a legal and/or moral right to an explanation if  ``choices are made which significantly affect us but which we do not understand''. 
If such a general right exists for a given context, regardless of whether the decision maker was a machine or a human, then we can argue that the replacement of the human decision-maker by an AI must not undermine the right to an explanation, the automated process ``inherits'' it from the manual, human decision maker.

\subsection{Explainability and the AI Act}\label{sec:background:aia}

In the GDPR, it is the right to contestation that is the end to achieve, in part via the means of transparency, opening the transparency gap.
However, the uncertainty around interpretations renders this argument weak, because the motivations for XAI could be readily adapted to bridge the gap.
By contrast, the AI Act~\cite{europeancommissionProposalArtificialIntelligence2021} clarifies, broadens, and operationalises transparency requirements and their effects on the legal requirements for explainability.\footnote{The Act has undergone significant changes in the EU Parliament and Council. For consistency with prior work on the Act, we use the final proposal by the EU Commission released on 21 April 2021. However, where appropriate, we will mention relevant amendments to the original proposal.}
Article 1(c) declares that the Act lays down `harmonised \textit{transparency rules} for AI systems intended to interact with natural persons [\dots]' (emphasis added).
The AI Act is an ambitious proposal for an EU-wide regulation presenting a sweeping set of rules aimed at harmonising and standardising compliance requirements for AI systems.
It takes a proportional risk-based approach to defining the transparency requirements, where \textit{high-risk} systems, such as facial recognition and law enforcement systems, are subject to greater regulation than low-risk ones.\footnote{See Annex II and III of the Act for a full list of systems classified as high-risk.}

In the following sections, we describe the transparency and explainability requirements of the Act building on the work of Sovrano et al.~\cite{sovranoMetricsExplainabilityEuropean2022}.
We review their discussion of explainability requirements, expand their reading with further requirements, and give an updated view that includes recent revisions of the Act.
To start, the Act distinguishes between \textit{user-empowering} and \textit{compliance-oriented} transparency.

\subsubsection{User-Empowering Transparency}\label{sec:law:user-transparency}

Article 13(1) addresses user-empowering transparency directly, stating that `high-risk AI systems shall be designed and developed in such a way to ensure that their operation is sufficiently transparent'.
This is to `enable users to \textit{interpret} the system's output sufficiently' (emphasis added), not just to facilitate the correct use of the system.
The first concrete user-empowering requirement for explanations appears in Article 13(2) which introduces an ex-ante explainability -- created prior to running the system -- requirement in the form of instructions for use that are `concise, complete, correct, and clear'.
That this is an explainability requirement is confirmed in Article 13(3)(b) which requires the instructions for use to contain relevant information as regards `the characteristics, capabilities, and limitations of performance of the high-risk AI system'.
Recital 47 also makes it clear, that these are essential for when high-risk `AI systems [are] incomprehensible or too complex for natural persons'.

In addition, \textit{human oversight} is a core element of the Act, which creates further explainability requirements.
To codify this view, Article 52(1) states that users should be made aware that they are interacting with an AI system.
Furthermore, Article 14(1) stipulates that `high-risk AI systems shall be designed and developed in such a way [\dots], that they can be effectively overseen by natural persons'.
According to Article 14(4), these measures must enable people to `fully understand the capacities and limitations', and to `correctly interpret the high-risk AI's output'.
Even more importantly, Article 14(4)(c) explicitly addresses, among others, explainable AI which it refers to as `interpretation tools and methods'.
Therefore, these paragraphs seem to place ex-post explainability -- created after a decision was made -- requirements on high-risk AI systems.
Recitals 38--40 mention the cases of law enforcement, migration, and administration of justice, where human oversight needs to be ascertained, as biased decisions have particularly far-reaching effects in these applications.

\subsubsection{Compliance-Oriented Transparency}

The Act also places strong requirements on compliance-oriented transparency.
In particular, Articles 9 and 17 establish the requirements for a risk-management and quality-management system.
These systems place detailed transparency requirements, achieved through documentation, monitoring, and verification, on the providers of high-risk AI systems to guarantee compliance with the Regulation.
Article 11 expands on the \textit{technical documentation} requirements of providers, which need to be drawn up before the high-risk AI system is placed on the market.
Referring to Annex IV(2)(b)-(d), Article 11 requires that such documentation includes `the general logic of the AI system and of the algorithms; the key design choices [\dots]; [and] the main classification choices'.
These are clear requirements on the ex-ante explainability of high-risk AI.

As Article 29(4) states, `users shall monitor the operation of a high-risk AI system on the basis of the instructions for use'.
This means that compliance-oriented transparency requirements need to enable users to monitor the operation of the high-risk AI system, forming a crucial interaction of user-empowering and compliance-oriented transparency requirements~\cite{sovranoMetricsExplainabilityEuropean2022}.
Additionally, Article 12 requires \textit{record-keeping}, or logging, of the high-risk AI system's operation.
Relatedly, the version of the Act from the Czech presidency of the EU Council adds Article 13(3)(f) requiring an ex-post `description mechanism [\dots] that allows users to properly collect, store, and interpret the logs'~\cite{counciloftheeuropeanunionGeneralApproachComission2022}.
Finally, Recital 58 states that responsibility has to extend to the users to maintain the correct operation of high-risk AI systems.
Therefore, an accurate and relevant explanation of the system is essential, otherwise, the user would not be capable of handling the system correctly.

Finally, we note that the Act has also attracted a lot of criticism~\cite{vealeDemystifyingDraftEU2021}.
A detailed discussion of these criticisms is out of scope, but we mention two recurring issues that will merit further academic attention.
First, many of the proposed amendments to the Act would incorporate exclusion criteria to the documentation and transparency requirements due to a tension between intellectual property rights and transparency~\cite{europeanparliamentDraftOpinionCommittee2022,europeanparliamentDraftOpinionCommittee2022a}.
The Act is also criticised for posing overreaching compliance-oriented transparency requirements. 
To address this, amendments to Article 11 would allow for start-ups, small, and medium enterprises to fulfil the requirements in equivalent but less demanding forms~\cite{counciloftheeuropeanunionGeneralApproachComission2022}.
However, this might enable conformity avoidance due to self-regulation as discussed in~\cref{ssec:misalignment:ca}.

\section{Bridging the Transparency Gap}\label{sec:misalignment}

As we saw, there are major differences in how technology and the law understand transparency and this leads to the transparency gap.
XAI uses a specialist vocabulary that addresses the distinct but narrow challenge of algorithmic transparency as an end in itself to be achieved.
Regulations, by contrast, consider a wider view of transparency that views it as one of many means through which wider values are supported. 
Without a mutual conceptual understanding of what transparency is, expert discussions (e.g., XAI literature) would lack sufficient breadth to address all concerns of society, while courts and regulators might lack the ability to assess algorithmic transparency.
After all, it is up to these legal bodies, informed by expert discussions, to determine appropriate interpretations to the normative demands set by high-level laws, such as the Act, that are then given specificity in their interactions with reality.
In the following, we identify and discuss four critical axes -- informed by our previous discussions -- along which the transparency gap may be addressed.

\subsection{Scope of Transparency}\label{ssec:misalignment:transparency}

In the interpretation of the Act, transparency is an overarching property of the AI system achieved through the requirements detailed in~\cref{sec:background:aia}.
Yet as we saw, the transparency gap means that XAI focuses solely on the algorithmic properties of the decision-making process.
Both interpretations have different consequences on the actual design of transparent AI.

In the Act, transparency is required to \textit{an appropriate level}, but no distinction is made on what exactly is appropriate for different applications or tasks. 
One might imagine that different levels of transparency would be required for different high-risk AI systems, but no greater level of detail is given in the Act.
A proportional approach to transparency could be emphasised by focusing on stakeholders similarly to how XAI might address stakeholders (cf., \cref{sec:xai:stakeholder}).
Nevertheless, a more feasible interpretation would at least require an understanding of the limitations of the systems concerning its ``\textit{intended purpose}'', a term used throughout the Act.
This will also allow the user to build an appropriate level of trust in the system, rather than over- or under-relying on it.
Additionally, as discussed in section \ref{sec:law:user-transparency}, Article 14(4) places lofty requirements on the understanding that human users are required to have.
It is worth questioning whether such aims are possible: a full understanding of capacities and limitations is surely an impossible task for black-box models, while the task of correctly interpreting a system's output is complicated by what is legally meant by an \textit{output}, discussed in \cref{ssec:misalignment:definitions}.

In contrast, our discussion of terminology in~\cref{ssec:xai:output} showed that the interpretation of transparency-related concepts in XAI is often too specific and technology-focused.
It often ignores societal and cultural concerns, making XAI less appealing as a solution for transparency, especially because transparency should be understood not as an end in itself, but as a means to achieve a range of important, but also heterogeneous and potentially conflicting, social goods.
While social and human-centred XAI~\cite{millerExplanationArtificialIntelligence2019} have long been trying to bridge the transparency gap, the uptake of these methods has been slowed down due to brittle conceptual frameworks~\cite{millerExplainableAIDead2023}, quickly changing external requirements~\cite{aliExplainableArtificialIntelligence2023}, and the difficulty of subjective evaluation~\cite{rongHumancenteredExplainableAI2022}.
Besides addressing these issues, the Act will in many ways affect how the XAI systems of the future are designed, and the field should look towards the Act to find inspiration for conceptual and requirements-related clarity, not least to bridge the transparency gap.

We also suggest a hierarchical approach to XAI, which enables targeted explanations addressing the right ``cognitive holes'' of humans based on risk levels.
On the lower levels of this hierarchy, for low-risk AI, we would have fully automated explanations from purely numerical to higher-level conceptual explanations.
As a threshold is crossed, reaching high-risk systems, we would increasingly introduce human interaction to guide the output generation of XAI systems, e.g., via dialogue systems.
This restores human agency by allowing people to intervene or contest decisions in extraordinary circumstances.

\subsection{Uncertainty in Legal Definitions}\label{ssec:misalignment:definitions}

Legal definitions in the Act leave room for flexible interpretations which have significant effects on the interpretation of the regulations and the utility of XAI for transparency.
This is expected on some level given the high-level nature of the Act, but XAI and the transparency gap significantly complicates the applicability of these definitions.
We illustrate this by focusing on the crucial concepts of \textit{AI system} and its \textit{output}, comparing how the Act and XAI approach them.

The Act's definition of an AI system is a much-debated and amended point~\cite{counciloftheeuropeanunionGeneralApproachComission2022,europeanparliamentDraftOpinionCommittee2022,europeanparliamentDraftOpinionCommittee2022a}.
Originally, Article (3)(1) defined AI systems in terms of a list of technologies, but this was later revised in the European Parliament~\cite{europeanparliamentDRAFTCompromiseAmendments2023} to reflect the definition by the Organisation for Economic Co-operation and Development (OECD)~\cite{oecdRecommendationCouncilArtificial2019}: ``a machine-based system that can, for a given set of human-defined objectives, make predictions, recommendations, or decisions influencing real or virtual environments [\dots] with varying levels of autonomy.''

The final definition of AI has wide-ranging consequences on how transparency is achieved.
Many XAI tools rely on practices that fall under the currently proposed definition of AI\@.
Does then the XAI tool qualify as the same or as a different AI system separate from the underlying AI algorithm which they explain?
We argue that the XAI tools should \textit{not} be treated separately from the AI system they explain.
This is a natural view to take as we argue for bridging the transparency gap, and in this sense, we argue against the general use of model-agnostic methods. 
XAI tools will need to be calibrated to work well with not just the AI system but its entire ecosystem.
Taking the XAI system out of context by assessing it as a separate entity will inevitably lead to erroneous conclusions.
Regulators need to clarify the legal status of XAI under the Act, making it clear that XAI systems should be assessed as part of the overall AI system, and XAI should focus on developing task-specific tools that retain domain knowledge.

In addition, the definition of an \textit{output} of an AI system is also critical, as transparency requirements depend in part on what needs to be explained.
In defining AI systems, the Act exemplifies outputs with the terms `content, predictions, recommendations, or decisions'. 
However, there is no explicit definition of output.
This approach is insufficient as XAI tools differentiate between internal (e.g., feature weights) and external outputs (e.g., a classification) of the AI system.
While the user cannot act directly on internal outputs, XAI tools -- especially ante-hoc systems seen in~\cref{sec:xai:methods} -- leverage them to produce an explanation.
To bridge the transparency gap, it is important for both regulators and XAI to define the outputs they work with since applying the same requirements to different types of outputs could result in conflicting technologies and regulations.

\subsection{Conformity Assessment}\label{ssec:misalignment:ca}

Another area where addressing the transparency gap is important is the methods by which AI providers are required to achieve conformity to the Act's transparency requirements. 
Documentation required by Article 11 of the Act is expected to demonstrate conformity with the regulation, and Article 19 requires the providers of high-risk systems to undergo an assessment of that conformity, as outlined in Article 43. 
Moreover, Article 17 also requires a comprehensive quality management system to be in place to ensure conformity during the entire lifecycle of the AI system.

However, XAI tools introduce further complexities to the AI system which affect their conformity under the Act, often without considering the wider effects due to the transparency gap.
We suggest that regulators and XAI should work towards a unified assessment scheme, where the AI system and XAI tool are considered as one unit so that the new complexities introduced by XAI would not go unnoticed or be construed as the inherent capabilities of the underlying AI system.
Legal requirements should thus clarify how the assessment of XAI tools is carried out, focusing specifically on the improved capabilities of the AI system by virtue of the XAI system, while scientists should measure the socio-technical effects of XAI tools via the involvement of human participants from various stakeholder groups.

Interestingly, the Act does not address how the actual effects of explanations on users should be accounted for.
An incorrect explanation is arguably more damaging than no explanation at all, thus, explanations should be subject to stringent quality control and conformity assessment too.
In bridging the transparency gap, we can use explainability fact sheets that provide a comprehensive checklist for the assessment of the correctness of XAI methods~\cite{sokolExplainabilityFactSheets2020}.

The Act, as worded, cannot provide software developers with enough detail to lead to actionable design decisions due to its high-level nature.
In some cases, e.g., medical devices, \textit{Notified Bodies} check the conformity assessment of the developers, in many others, self-certification is sufficient. 
The combination of a limited role for Notified Bodies and the lack of detail in the Act gives industrial standards a particularly prominent role.
Though the Act does not mandate that developers adhere to industrial standards, this may be a ready way to narrow the transparency gap. 
While developers could interpret the requirements of the Act on their own, and decide how to implement transparency requirements, in this case, the risk of misreading the law rests with them.
If, by contrast, they adhere to any future standards developed by CEN (European Committee for Standardisation) and CENELEC (European Committee for Electrotechnical Standardisation), they are protected by a ``presumption of conformity''. 

As constituted, however, these standard-setting bodies lack the expertise and remit to consider for instance Human Rights implications of their standards. 
We discussed at our outset how transparency speaks to a whole range of important human rights that a badly designed AI system may impact: from the right to bodily integrity to the right to non-discriminatory treatment to the right to privacy. 
We have also seen how multifaceted the concept of transparency is.

Standard-setting bodies, because of the influence industry plays in them, are likely to emphasise compliance-oriented transparency over user-empowering transparency. 
Here, the Act's proposal may be much less efficient in protecting basic rights than, for example, the envisaged UK regulatory framework for AI~\cite{dorriesEstablishingProinnovationApproach2022}.
Unlike the top-down EU approach that may result in an attempt to define transparency for a whole range of disparate applications, the UK approach is emphasising domain-specific regulators. 
These regulators not only often have relevant legal expertise, and a statutory duty to consider human rights implications, but they are also better placed to understand the different roles transparency plays in their respective fields. 

Another problem arises when new models of AI systems are developed and released. 
Article 43(b) mandates recertification in the event that the system is substantially modified -- but what constitutes a substantial modification? 
For our context, in particular, do changes to the XAI tool count? 
A possible scenario here is a system that delivers generally correct results but has an XAI component that is increasingly not state-of-the-art.
If this component is updated, does the entire system need recertification?
After all, \textit{ex hypothesi} the substantial results have not changed, and the AI system  still conforms to industry-standards. 
If the answer is yes, then this might be a deterrent to upgrading the XAI component of a certified system. 

It is crucial for solving the transparency gap, that more of academia is involved in the work of not just standard-setting bodies but other regulatory bodies, to enable the work of these institutions and to understand the wider values that are at stake in misreading transparency.

\subsection{Explainable Data}\label{ssec:misalignment:data}

So far most of our discussion in XAI has focused on the algorithmic aspect of transparency.
However, data-related requirements and the subsequent design choices have clear transparency-related implications and these considerations are strongly regulated in the Act.
In particular, Article 10 and Recital 44 of the Act lays down comprehensive requirements for \textit{training}, \textit{validation}, and \textit{testing data} used with AI systems, addressing, either directly or indirectly, many of the Human Rights implications we eluded to in the previous section.

However, the Act misses a crucial technological aspect in Article 10 by constraining its scope to training, validation, and testing data.
Datasets are indeed crucial for data-driven AI systems, however, methods such as planning and reinforcement learning do not rely on the same techniques as supervised learning from where these terms originate~\cite{bishopPatternRecognitionMachine2006}.
It is unclear what requirements should be fulfilled for XAI systems that rely on the input data of such systems to generate their explanations.
More comprehensive coverage of data types is necessary to cover a wider range of algorithms.

Further indicative of the transparency gap is that XAI has a long way to go in addressing data-oriented concerns.
Automated \textit{data explanation} techniques could examine the effects of inherent properties of the input dataset on the AI system, for example by purposefully biasing, distorting, or introducing irrelevant samples to the dataset.
Such methods could identify issues with data entanglement as well -- a result of mixing multiple sources of data. 
Data explanations may uncover problems with the dataset and the system itself and, at the same time, conformity to Article 10(2) of the Act could be demonstrated.

Another promising direction in practical data governance is data documentation.
Gebru et al.~\cite{gebruDatasheetsDatasets2021} introduce the ``datasheet for datasets'', which is a careful way to trace the motivation, creation, and qualities of a dataset.
Measures like this would improve AI transparency in a way that is compatible with the Act, bridging the transparency gap.

\section{Conclusion}\label{sec:conclusion}
The transparency gap presents the very real risk that the differing views of law and computer science on transparency will prevent the Act and XAI from achieving a beneficial impact. 
This could happen in a number of ways: one is identifying the wider legal concept of transparency with the much narrower computer science concept, allowing the technological discourse to replace the socio-legal one. 
This would leave a range of wider transparency harms unaddressed.

The other danger is that legislators overestimate the capabilities of XAI and as a result overload the regulation with unachievable, highly detailed design prescriptions that will often opt to deliver benefits for users and affected third parties, not necessarily because they fail in generating transparency, but because the legal framework does not then give the affected parties the tools to act on what they learned.

In the first scenario, we lower our expectations of the law too much, in the second we raise our expectations of computer science too high. 
Instead of computer science colonising law, or vice versa, we suggested four approaches that respect the internal logic of the two systems and ask XAI not to “solve” the problem of legal transparency, but to understand how, given the internal logic of the law, it can develop new tools and approaches to facilitate transparency while staying true to its foundations at the same time.

\ack This work was supported in part by the UKRI Centre for Doctoral Training in Natural Language Processing (grant EP/S022481/1) and the UKRI Trustworthy Autonomous Systems Node in Governance and Regulation (grant EP/V026607/1).

\bibliography{refs}
\end{document}